\documentclass[letterpaper, 10 pt, conference]{ieeeconf}  

\IEEEoverridecommandlockouts                              

\usepackage{graphics} 
\usepackage{epsfig} 
\usepackage{mathptmx} 
\usepackage{times} 
\usepackage{amsmath} 
\usepackage{amssymb}  
\usepackage{hyperref}
\usepackage{float}
\usepackage{booktabs}
\usepackage{caption}

\title{\LARGE \bf NavGen: Visual Generative Models as a Scalable Data Engine \\for Embodied 3D Navigation}

\author{%
      Xijie Huang\textsuperscript{1,2},
      Yongyang Wan\textsuperscript{2},
      Chengbin Dong\textsuperscript{2},
      Zimo Ding\textsuperscript{2},
      Mo Zhu\textsuperscript{1,2},
      Yijin Wang\textsuperscript{1,2},\\
      Zhiyang Liu\textsuperscript{2},
      Fei Gao\textsuperscript{1,2},
      Yuze Wu\textsuperscript{1,2,\textdagger},
      and Xin Zhou\textsuperscript{2}\\
      \thanks{\textsuperscript{1} Zhejiang University.}%
      \thanks{\textsuperscript{2} Differential Robotics.}%
      \thanks{\textsuperscript{\textdagger} Corresponding author: \texttt{wuyuze000@zju.edu.cn}}%
      Project page: https://xinjiu612.github.io/NavGen
}

\begin{document}
\bstctlcite{BSTcontrol}

\maketitle
\thispagestyle{empty}
\pagestyle{empty}


\begin{figure*}[t]
  \centering
  \includegraphics[width=1.0\textwidth, height=8.5cm]{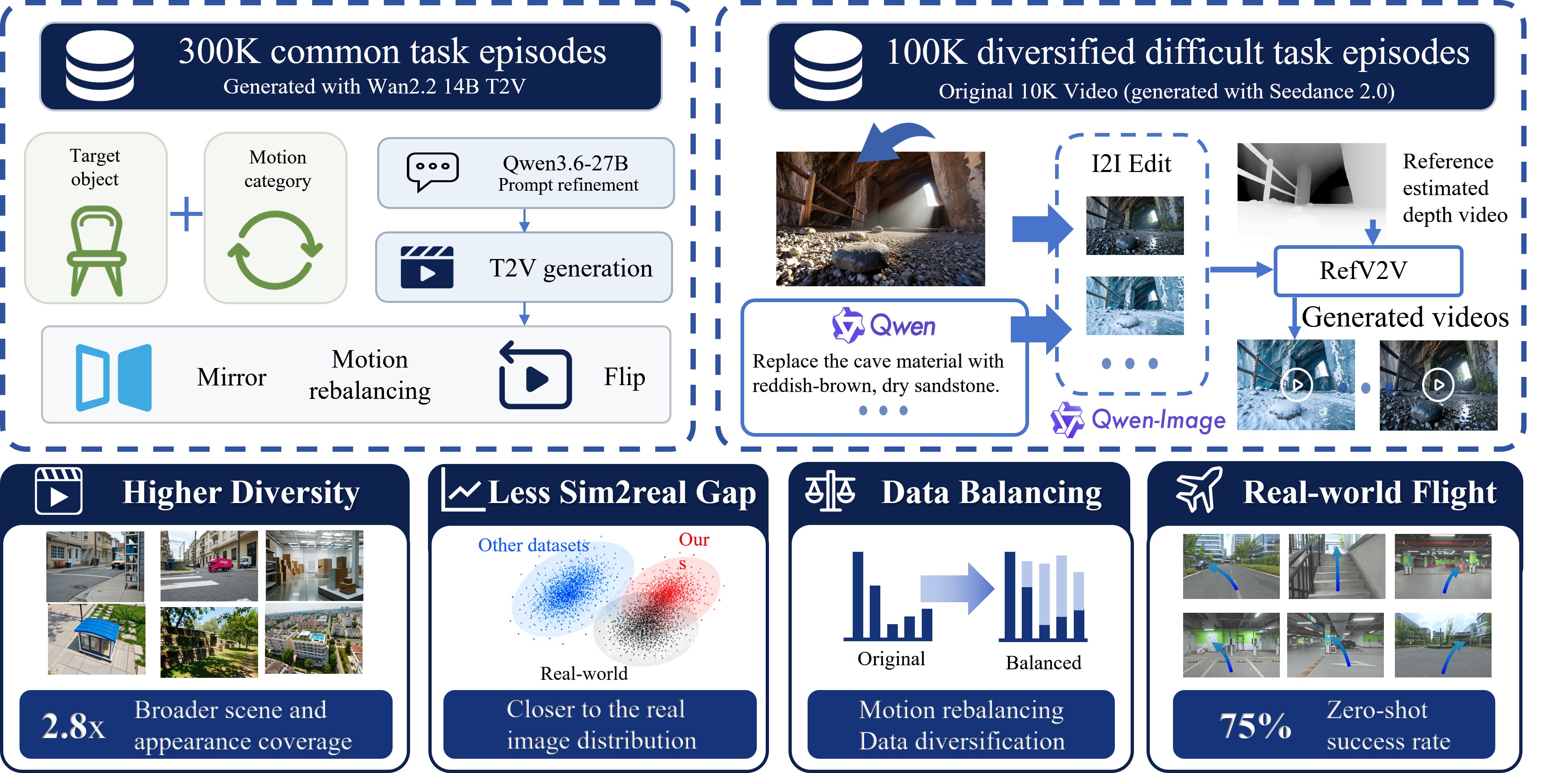}

  \caption{\textbf{Overview of the NavGen data generation pipeline.} The upper left part uses the open-source model to generate 300K common task samples. The upper right part uses the more powerful but closed-source model to generate 10K long-tail task samples with a style-diversification pipeline to scale up to 100K.}
  \label{fig:toutu}
\end{figure*}

\begin{abstract}

General-purpose robot models increasingly rely on large and diverse datasets. For embodied 3D navigation, however, existing data sources face a fundamental trade-off: simulated data can be generated at scale but often suffer from the visual sim-to-real gap, whereas real-world flight data provide realistic observations but are costly to collect. This paper studies another direction: the use of high-fidelity visual generative models as scalable data engines for embodied 3D navigation. We introduce NavGen, a text-to-video data generation pipeline that produces diverse vision-language navigation (VLN) episodes across indoor and outdoor scenes. We also propose a style-diversification method that scales up long-tail data that are difficult and costly to collect. The resulting dataset contains approximately 400K navigation episodes. We evaluate our dataset against existing UAV navigation datasets across multiple metrics, and find that the model trained on our data generally improves with scale, outperforming those trained on existing datasets.  To validate real-world transferability, we deploy the trained model in world-action-model paradigm to real-world flying experiments. The final model achieves a 75\% success rate across different navigation tasks and environments.

\end{abstract}

\section{INTRODUCTION}


Embodied 3D navigation requires an agent to interpret language and visual observations while producing temporally coherent motion in complex environments. Its generalization, however, is ultimately constrained by the range of environments and behaviors represented in the training data. Building such data at scale remains difficult because navigation is not merely a problem of collecting more trajectories: the data must also be visually realistic, behaviorally rich, and diverse in scenes, objects, and tasks.

But existing data sources force us to trade one of these properties for another. Simulation platforms such as AirSim \cite{airsim}, Matterport3D \cite{Matterport3D}, and Unreal Engine can produce many trajectories, but their appearance and discrete action abstractions introduce visual and dynamic gaps with respect to the physical world \cite{traveluav,indooruav,openfly}. Real-world flight data provide more realistic observations, but collecting them is costly and difficult to scale \cite{uavflow}; challenging navigation tasks can also put the aircraft at risk of crashing. OpenFly \cite{openfly} proposes an automated pipeline that can continuously collect trajectories at scale. However, when the underlying scenes, objects, and task configurations are limited, simply increasing the number of trajectories can result in highly homogeneous data, providing limited additional supervision for improving the model's generalization ability. Overall, existing pipelines do not yet provide a practical way to collect navigation data that is simultaneously realistic, diverse, and scalable.

Given this tension, we ask whether visual generative models can play a broader role in embodied intelligence: not only serving as policies, but acting as scalable engines for producing navigational experience. Trained on massive real-world data, these models can synthesize high-fidelity indoor and outdoor scenes, vary object appearances and visual styles, and express camera motion through natural-language or visual conditions.

In this paper, we present NavGen, a scalable data engine that composes visual generative models to produce navigation data. We show that navigation performance consistently improves as the amount of generated data grows. We first use the open-source, low-cost Wan2.2 T2V 14B \cite{wan} with a diverse text-to-video prompting strategy to produce 300K episodes covering common embodied navigation tasks. For long-tail maneuvers beyond the open-source model's capability, such as orbiting a target or passing through narrow spaces, we instead use a more powerful closed-source model \cite{seedance2}, to generate 10K high-quality samples. To enrich these long-tail data, we scale these samples to 100K through a style-diversification module that combines an image-editing model \cite{qwen_image} with a reference-conditioned video-to-video model \cite{ltx}. All episodes then undergo motion rebalancing, video filtering, and prompt re-annotation. The resulting 400K-episode dataset provides scale, diversity, and long-tail coverage at once.  

As world action models have emerged as a promising paradigm for embodied AI \cite{lingbotva}, we adopt the video-prediction and action-decoding pipeline of NavDreamer \cite{navdreamer} to evaluate how different training datasets affect navigation performance. Specifically, we train the lightweight Wan2.2 5B video model separately on our dataset and existing embodied 3D navigation datasets. Extensive comparisons show that our dataset is 2.8$\times$ more diverse than the strongest existing dataset, has the smallest sim-to-real gap among non-real datasets, and yields a 17\% higher navigation progress score than the best baseline. Furthermore, we observe that increasing the scale of our dataset consistently improves the model's navigation performance. We finally build a low-latency inference system for real-world deployment, and the deployed UAV completes diverse tasks in unseen environments with a 75\% zero-shot success rate.
  
Our contributions are:                                          
\begin{itemize}                                           
\item NavGen, a scalable data engine that composes visual generative models (with motion rebalancing, data diversification, video filtering and re-annotation) into a 400K-episode navigation dataset providing scale, diversity, and long-tail coverage.                   
\item A controlled benchmark quantifying the data's value: 2.8$\times$ higher diversity, the smallest sim-to-real gap among non-real datasets, and a 17\% higher navigation progress score over the best baseline.                                         
\item Real-world validation of the data engine: a 47-FPS video generation system deploying the trained model on a physical UAV with a 75\% zero-shot success rate in unseen environments.                         
\item Our dataset, model weights, and code will be made publicly available upon acceptance.

\end{itemize}

\section{RELATED WORK}

\subsection{Embodied 3D Navigation Datasets}

Embodied 3D navigation has benefited from large simulated environments such as Matterport3D, AirSim, and Unreal Engine \cite{Matterport3D,airsim}. TravelUAV \cite{traveluav} collects trajectories in AirSim and Unreal Engine environments, while IndoorUAV \cite{indooruav} extends this direction to indoor navigation with human-operated trajectories in Matterport3D assets. These datasets rely on simulation or manual piloting, which limits their realism, scalability, and diversity.

OpenFly \cite{openfly} proposes an automated collection pipeline using AirSim, GTA5, 3DGS \cite{3dgs} scenes, and Google Earth, reducing human operation cost but still depending on simulator appearance and manually designed task rules. UAV-Flow \cite{uavflow} instead collects real-world UAV trajectories, but its scenes are concentrated in limited campus environments. VLA-AN \cite{vla_an} uses navigation data collected in 3DGS environments to mitigate the sim-to-real gap. FlyMirage \cite{flymirage} further explores automated collection in generated 3DGS scenes with a world generation model, but it still requires explicit task design and is affected by blur and artifacts of 3DGS reconstructions.

\subsection{Visual Generative Models for Embodied AI}

Recent progress in image and video generation models has motivated their use as data sources for embodied AI. Early works mainly focus on data augmentation: ROSIE \cite{rosie} uses a text-to-image diffusion model to paint new objects, backgrounds, and distractors; RoboTransfer extends appearance editing to video and multi-view settings \cite{robotransfer}; and RoboCurate \cite{robocurate} couples generation with verification by filtering generated trajectories through simulation replay. These methods improve visual diversity but remain tied to the initial demonstration distribution.

Another line of work synthesizes complete embodied trajectories. DreamGen \cite{dreamgen} adapts image-to-video models to robot embodiments, but it requires manually provided high-quality and diverse initial frames. X-Humanoid \cite{xhunmanoid} and Lucid-XR \cite{lucidxr} generate large-scale robot-centric visual data for humanoid or manipulation policies. Although these works demonstrate the potential of visual generative models for embodied AI, a systematic framework for generating diverse and high-quality embodied datasets through different generation strategies is still lacking. Moreover, how to transfer this data-generation paradigm to embodied navigation remains underexplored.

\section{METHOD}

We focus on building a balanced data-generation pipeline. In Section~\ref{diverse_prompt}, we describe how we maintain sample diversity at the 300K scale. In Section~\ref{motion_rebalancing_style_diversification}, we propose a motion rebalancing strategy to correct the distribution bias of video generation models, and scale up 10K long-tail samples from a closed-source model to 100K through style diversification. In Section~\ref{video_description}, we use Qwen3.6-27B to re-annotate the generated videos as training prompts. Finally, we design a low-latency inference system that achieves 47 FPS and introduce our training system design in Section~\ref{training}.

\subsection{Diverse Embodied 3D Navigation Descriptions for Text-to-Video Generation}

\label{diverse_prompt}
Automatically generating diverse navigation task descriptions for text-to-video generation is a core component of our pipeline. A straightforward solution is to use a vision-language model (VLM) to generate prompts. However, when the
dataset size reaches hundreds of thousands of samples, VLM responses tend to become increasingly homogeneous. To ensure sufficient diversity in our dataset, we start from an object dataset \cite{cap3d} containing 1M objects, filter out uncommon objects, and obtain 300K object targets.

To keep the motion distribution balanced, we assign the 300K samples to a predefined set of motion categories. Given the pre-designed motion prompt and the target object, we use Qwen3.6-27B as an embodied 3D navigation expert to imagine detailed navigation events around the object, including the environment style, background objects, spatial layout, and target-related context. This produces an automatically generated embodied 3D navigation text dataset, while ensuring object-level diversity across different samples, as shown in Fig.~\ref{fig:toutu}.

With the generated navigation descriptions, we use the step-distilled Wan2.2 14B T2V \cite{lightx2v} model to generate videos with four diffusion steps. Each sample takes approximately 36 seconds to generate on an A100 GPU.

\subsection{Motion Rebalancing and Style Diversification}
\label{motion_rebalancing_style_diversification}

\subsubsection{Motion Rebalancing}
We observe that video generation models may not strictly follow the designed prompts, leading to an imbalanced motion distribution that introduces bias into the trained model. To address this, we propose a motion rebalancing strategy.

Specifically, we first use Pi3 \cite{pi3} to decode the overall motion direction of each generated video and assign a motion label to every sample. We then identify the imbalanced motion categories and divide them into two major types: left-right imbalance and forward-backward imbalance. For left-right imbalance, we use horizontal mirroring to rebalance the distribution. For forward-backward imbalance, we use temporal reversal to increase the underrepresented motion direction. In this way, we improve the balance of the generated motion
distribution without requiring additional video generation.

\subsubsection{Style Diversification}

For challenging long-tail tasks, we use the Seedance API to generate 10K high-quality samples, including difficult behaviors such as passing through narrow spaces and orbiting around targets. These tasks require stronger temporal consistency and more accurate motion control from
the video generation model. To scale up these samples, we extract the first frame of each video and use Qwen3.6-27B to generate style-diversification descriptions that preserve the main object shape while changing the surrounding objects, environment style, and scene appearance, as shown in Fig.~\ref{fig:toutu}. We
then use Qwen-Image-Edit \cite{qwen_image} to edit the first frames according to the generated diversification descriptions.

Finally, we use the LTX-2.3 \cite{ltx} ICLoraPipeline to generate new videos conditioned on the original reference videos. Before this step, the original videos are converted into depth videos using MoGe-2 \cite{moge}, which provides structural guidance for the generation
process. This pipeline allows us to automatically scale the original 10K long-tail samples to 100K samples, and we validate its effectiveness in our experiments.

\subsection{Video Filtering and Description}
\label{video_description}

Since video generation models may occasionally hallucinate and produce content that violates consistency, we filter the generated videos with an automatic evaluation pipeline (Section~\ref{nav_evaluation}). Specifically, we discard samples whose VBench~\cite{vbench} score falls below a predefined threshold or whose navigation grade is below L2, and regenerate them from newly sampled navigation descriptions. This step removes low-quality and inconsistent videos and improves the overall reliability of the dataset.

Since the generated videos may not perfectly follow the original detailed prompts, generating accurate video-level event descriptions is essential for training. We use Qwen3.6-27B to describe each generated video by uniformly downsampling it to 11 frames and using them as input. Because VLMs often have limited spatial-direction understanding, we additionally provide the Pi3-decoded overall motion direction as a motion cue for VLMs, which improves the quality and consistency of the generated descriptions.

\begin{table}[h]
\centering
\caption{Inference latency breakdown of the proposed system.
T5, Enc., DiT, and Dec. denote text-encoding, VAE-encoding,
diffusion-transformer, and VAE-decoding latency, respectively.}
\label{tab:time_compute}
\setlength{\tabcolsep}{3pt}
\renewcommand{\arraystretch}{1.15}
\resizebox{\columnwidth}{!}{%
\begin{tabular}{lccccc|c}
\toprule
\textbf{Variant} & \textbf{T5} & \textbf{Enc.} & \textbf{DiT} & \textbf{Dec.} & \textbf{Total} & \textbf{Speedup} \\
 & (s) & (s) & (s) & (s) & (s) & ($\times$) \\
\midrule
Original                            & 0.116 & 0.189 & 156.583 & 13.520 & 171.568 & 1.00$\times$ \\
Sparse                              & 0.165 & 0.057 & 9.585   & 1.421  & 12.250  & 14.00$\times$ \\
\;+ Distilled                       & 0.058 & 0.051 & 0.553   & 1.040  & 1.751   & 97.98$\times$ \\
\;+ Distilled + TAE                 & \textbf{0.058} & \textbf{0.049} & \textbf{0.510} & \textbf{0.055} & \textbf{0.702} & \textbf{244.40$\times$} \\
\bottomrule
\end{tabular}}
\end{table}

\subsection{Training System Design}
To evaluate our dataset and test its sim-to-real transferability, we adopt the video-prediction and action-decoding pipeline of NavDreamer~\cite{navdreamer} with the lightweight Wan2.2 5B model as the base model. For real-world deployment, we first use a bidirectional long-video generation paradigm and train the model on our dataset through supervised fine-tuning (Sections~\ref{long_video} and \ref{sft}). We then accelerate inference with a distillation pipeline and a lightweight VAE (Section~\ref{distill}).

\label{training}

\subsubsection{Bidirectional Long-Video Generation via Clean-Prefix History Conditioning}
\label{long_video}
\begin{figure}[h]
  \centering
  \includegraphics[width=1.0 \linewidth]{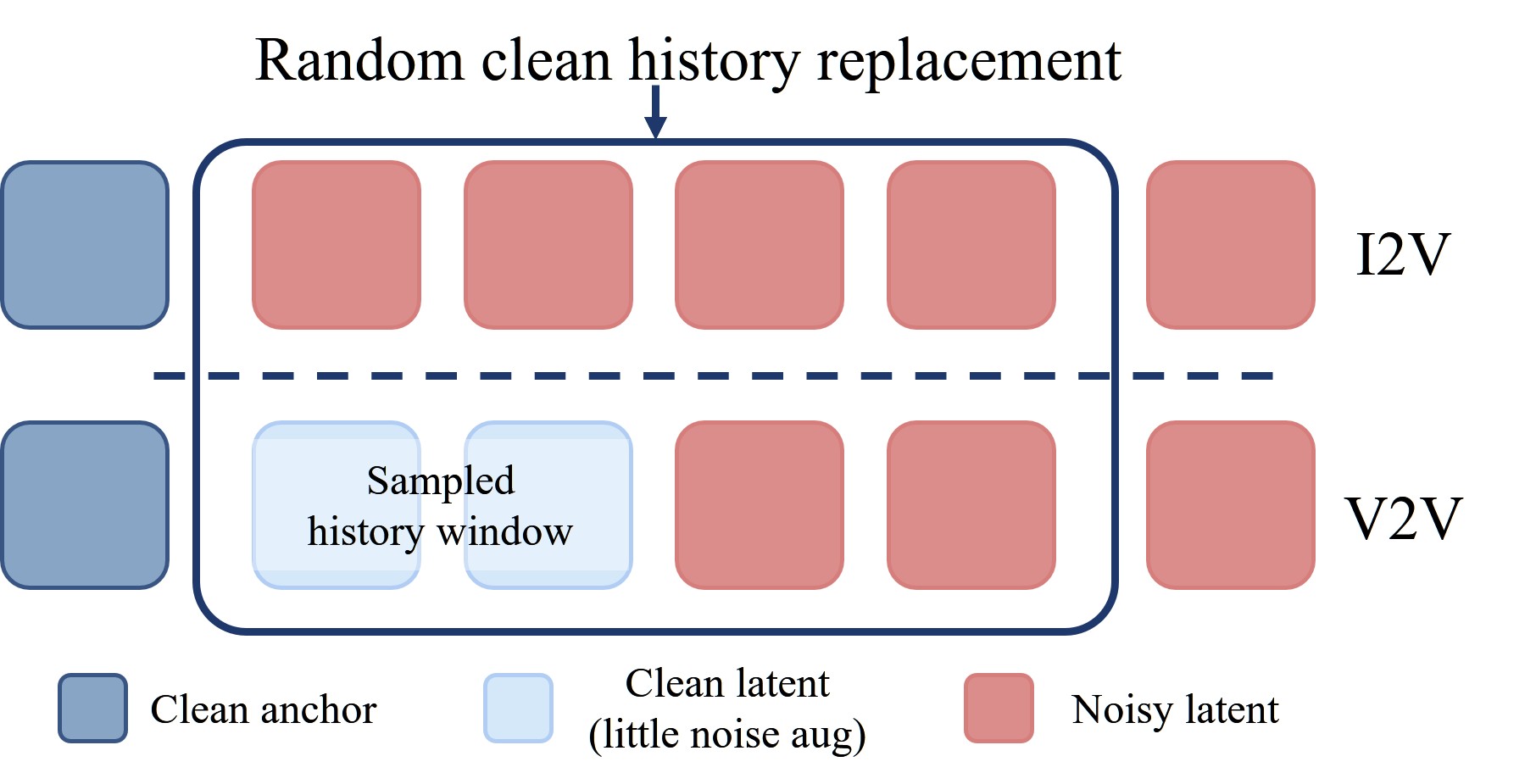}
  \caption{\textbf{Bidirectional long-video generation via clean-prefix history conditioning.}}
  \label{fig:long_video}
\end{figure}

Although the autoregressive paradigm supports unbounded continuous video generation, we find it yields lower task success rates than the bidirectional paradigm. However, most navigation tasks cannot be completed within a single clip, and conditioning each clip independently on only the current anchor frame causes inconsistent motion dynamics across successive clips. We address both issues with a unified image-to-video (I2V) and video-to-video (V2V) training requiring no architectural change as shown in Fig.~\ref{fig:long_video}.

During training, the first $k$ latent tokens are replaced with their clean ground-truth observations, where $k \sim \mathrm{Uniform}\{1,\ldots,N\}$ is sampled each step independently. Those latent tokens are excluded from the loss, and the remaining tokens are denoised under standard full bidirectional attention. Setting $k{=}1$ recovers the I2V objective, while $k{>}1$ trains the model to continue from $k{-}1$ clean history latents, i.e., V2V. A small noise augmentation with probability $p_{\text{aug}}$ is applied to history frames to prevent over-reliance on a perfectly clean prefix. Because $k$ is randomized, a single model covers both the I2V and V2V.

\subsubsection{Supervised Fine-Tuning}
\label{sft}

Embodied 3D navigation is typically a long-horizon task, where adjacent frames are highly redundant and high-resolution generation is not always necessary. We find that a resolution of $640 \times 384$ is sufficient for most navigation behaviors, while substantially reducing the computational cost compared with $1280 \times 720$ generation. Furthermore, we find that a video model trained on sparse frames can efficiently represent long-horizon events using fewer frames. Therefore, we first train a Wan2.2-5B TI2V model at $640 \times 384$ resolution with 33 frames. The training videos are resized and temporally downsampled from the 81-frame outputs generated by the Wan2.2-14B T2V model. This design preserves the main task completion process while reducing both spatial and temporal redundancy. The total generation time is reduced from 171.57s to 12.25s.

We first perform supervised fine-tuning (SFT) to adapt the student model to the generated navigation data. Following the flow-matching formulation, we construct the noisy latent $z_t = (1-t) z_0 + t \epsilon$ with the flow target $u_t = \epsilon - z_0$, where $z_0$ is the clean video latent, $c$ is the text condition, $t \in [0,1]$ is the diffusion timestep, and $\epsilon \sim \mathcal{N}(0,I)$ is Gaussian noise. The SFT objective is:
\begin{equation}
\mathcal{L}_{\mathrm{SFT}}
=
\mathbb{E}_{z_0,c,t,\epsilon}
\left[
\left\|
f_{\theta}(z_t,t,c) - u_t
\right\|_2^2
\right]
\end{equation}
where $f_{\theta}$ is the student denoising network. This stage provides a strong initialization before step distillation.

\subsubsection{Step Distillation}

\label{distill}
After SFT, we reduce the number of sampling steps through continuous-time consistency distillation. We warm up the model for 2.5K steps using the simplified continuous-time consistency model objective (sCM) \cite{scm}. Under the TrigFlow parameterization $z_t = \cos(t) z_0 + \sin(t) \epsilon$ with $t \in [0, \pi/2]$, the sCM training gradient is:
\begin{equation}
\nabla_{\theta}\mathcal{L}_{\mathrm{sCM}}
=
\nabla_{\theta}
\mathbb{E}_{z_t,t,c}
\left[
w(t) f_{\theta}^{\top}(z_t,t,c)
\frac{d f_{\theta^-}(z_t,t,c)}{d t}
\right]
\end{equation}
where $f_{\theta}$ is the consistency function parameterized by the student network, $f_{\theta^-}$ is the stop-gradient target, and $w(t)$ is the timestep-dependent weighting function. In practice, this stage serves as a warmup to stabilize the student before stronger distribution-level distillation.

We then apply DMD2-style distribution matching distillation \cite{dmd2}. The student generator $G_{\theta}$ produces a video latent $\hat{z}_0 = G_{\theta}(z_T,c)$ from noise and condition $c$. The student is trained by matching its generated distribution to the teacher's via the velocity difference:
\begin{equation}
\small
\nabla_{\theta}\mathcal{L}_{\mathrm{DMD}}
=
\mathbb{E}_{t,\hat{z}_0,c,\epsilon}
\left[
w(t)
\left(
v_{\psi}(z_t,t,c)
-
v_{\phi}(z_t,t,c)
\right)
\partial_{\theta}\hat{z}_0
\right]
\end{equation}
where $v_{\psi}$ and $v_{\phi}$ are the teacher and fake score models (trained on student samples), respectively, and $z_t$ is the noised latent. The fake model is optimized via flow matching:
\begin{equation}
\mathcal{L}_{\mathrm{fake}}
=
\mathbb{E}_{\hat{z}_0,t,c,\epsilon}
\left[
\left\|
v_{\phi}(z_t,t,c) - (\epsilon - \hat{z}_0)
\right\|_2^2
\right]
\end{equation}
To improve visual quality, we add an adversarial loss. Following DMD2, the discriminator $D_{\eta}$ operates on diffused samples via the forward process $F(\cdot, t)$:
\begin{equation}
\footnotesize
\mathcal{L}_{D}
=
-
\mathbb{E}_{z_0,t}
\left[
\log D_{\eta}(F(z_0,t))
\right]
-
\mathbb{E}_{\hat{z}_0,t}
\left[
\log
\left(
1-D_{\eta}(F(\hat{z}_0,t))
\right)
\right]
\end{equation}
and the generator-side GAN loss is:
\begin{equation}
\mathcal{L}_{\mathrm{GAN}}
=
-
\mathbb{E}_{\hat{z}_0,t}
\left[
\log D_{\eta}(F(\hat{z}_0,t))
\right]
\end{equation}
The final student objective is $\mathcal{L}_{\mathrm{student}} = \mathcal{L}_{\mathrm{DMD}} + \lambda_{\mathrm{GAN}}\mathcal{L}_{\mathrm{GAN}}$. Both the student and fake models are initialized from the sCM-warmed checkpoint, while the teacher uses the SFT checkpoint.

As shown in Table~\ref{tab:time_compute}, the Sparse processing at a reduced resolution of $640\times384$ with 33 frames instead of the full $1280\times720$ with 121 frames yields a $14\times$ speedup. Step distillation then reduces the denoising process from 30 steps to 4 and removes classifier-free guidance (CFG), achieving a $17\times$ reduction in DiT (Diffusion Transformer) computation. At this point, the VAE becomes the bottleneck: encoding and decoding together need to take about 1 second, comparable to the entire 0.5-second DiT pass. We address this by replacing the standard VAE with a lightweight alternative~\cite{taehv}, reducing the combined encode and decode time to roughly 0.1 seconds. With these three optimizations, the full inference pipeline runs in 0.7 seconds per clip, corresponding to 47 FPS. Training hyperparameters are summarized in Table~\ref{tab:training_hyperparams}.

\begin{table}[h]
\centering
\scriptsize
\caption{Training hyperparameters across the three-stage pipeline.}
\label{tab:training_hyperparams}
\begin{tabular}{lccc}
\toprule
\textbf{Parameter}
  & \textbf{Stage 1: SFT}
  & \textbf{Stage 2: sCM}
  & \textbf{Stage 3: DMD2} \\
\midrule
LoRA rank              & 64      & 64      & 64      \\
LoRA alpha             & 64      & 64      & 64 \\
Global batch size      & 32      & 32      & 32      \\
Optimizer              & AdamW   & AdamW   & AdamW   \\
Generator LR           & 1e-4    & 1e-5    & 2e-5    \\
Critic LR              & --      & --    & 4e-6    \\
Training Steps         & 12500   & 2500    & 5000    \\
GPU Usage              & 16 NVIDIA A100 & 16 NVIDIA A100 & 16 NVIDIA A100 \\
\bottomrule
\end{tabular}
\end{table}

\begin{figure}[h]
  \centering
  \includegraphics[width=0.5\textwidth]
  {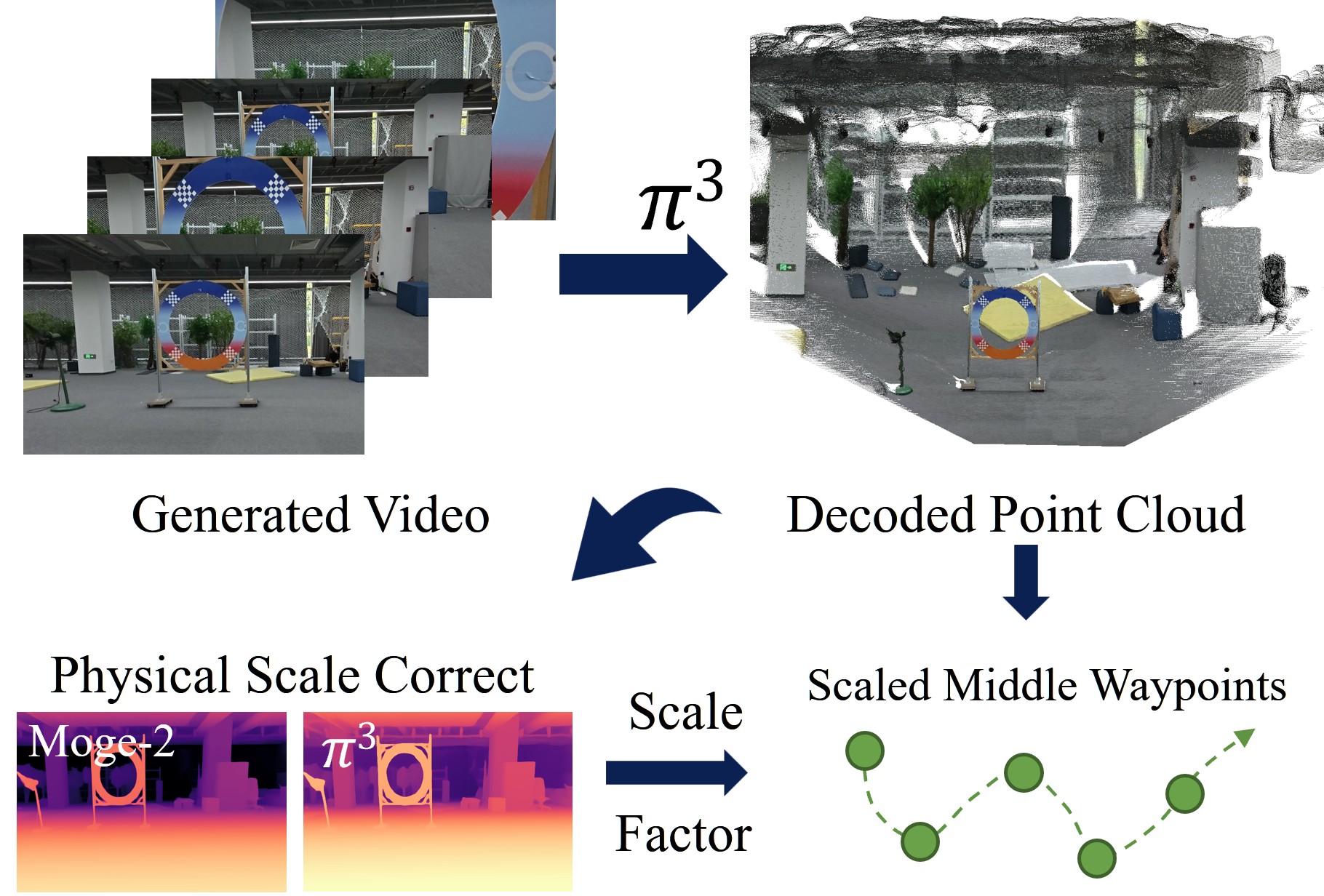}
  
  \caption{\textbf{Waypoints decoding from RGB frames.}}
  
  \label{fig:real_world}
  \vspace{-2mm}
\end{figure}

\section{EXPERIMENTS}

\subsection{Dataset Analysis}
To quantitatively analyze the motion characteristics of the 400K generated navigation samples, we employ an action-decoding pipeline \cite{navdreamer}, which is consistent with our real-world deployment. Given a generated RGB video, we first use Pi3 \cite{pi3} to estimate the camera trajectory and geometry as shown in Fig~\ref{fig:real_world}. However, Pi3 suffers from scale ambiguity, particularly in outdoor environments. To recover precise metric-scale trajectories, we introduce a metric scale correction module using MoGe-2 \cite{moge}. Specifically, we compute a scale factor by aligning the metric depth estimated by MoGe-2 with the depth rendered from the Pi3 point cloud at corresponding viewpoints. This scale factor is then applied to the Pi3 trajectory, yielding physically accurate motion sequences.

\begin{figure}[h]
  \centering
  \includegraphics[width=0.98\linewidth]{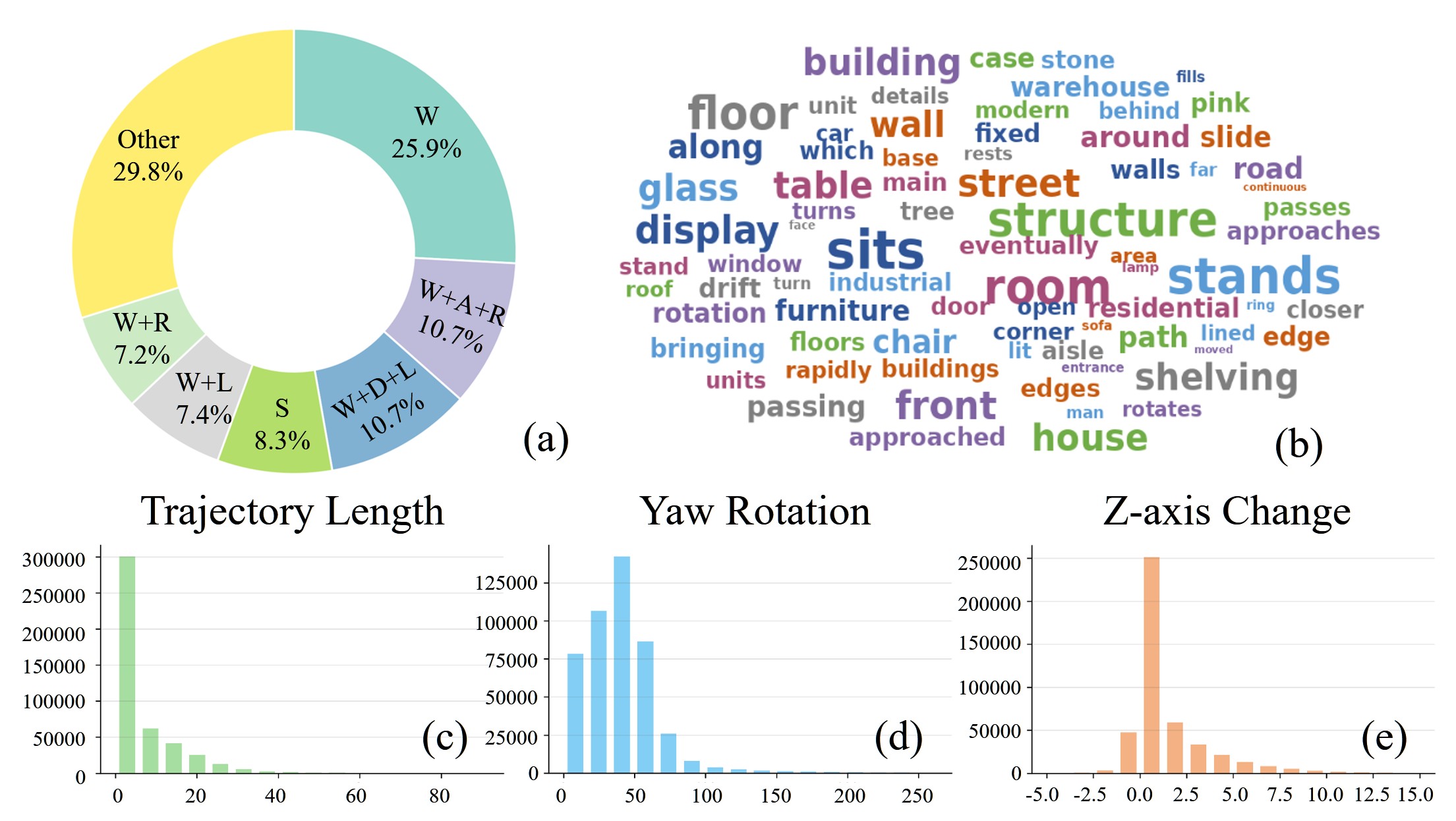}
  \caption{\textbf{Qualitative analysis of our dataset.} Figure (a) is the motion distribution. Figure (b) is the word cloud of the text description. Figure (c), (d), (e) are the analyses of the specific dimension.}
  \label{fig:analysis}
  \vspace{-2mm}
\end{figure}

Fig.~\ref{fig:analysis} summarizes the resulting dataset statistics. Unlike prior datasets that often describe navigation with discrete action labels, our video-generated trajectories provide continuous egocentric motion, making them more suitable for modeling fine-grained aerial behaviors. Since video generation models may occasionally deviate from the initially specified motion trends, the resulting dataset can become biased toward certain motion directions. Our motion rebalancing strategy can correct this distributional imbalance efficiently. The dataset is mainly composed of short-range trajectories, similar to UAV-Flow \cite{uavflow}, but still exhibits diverse yaw rotations and vertical movements, reflecting broad motion coverage across different navigation scenarios.

\begin{table}[h]
    \centering
  \caption{VBench grouped scores at the best-performing checkpoint per dataset (bold = best, underline = second best per row). Aesthetic: Aesthetic Quality; Smoothness: Motion Smoothness and Temporal Flickering; Sem.\ Cons.: Subject and Background Consistency; I2V Cons.: alignment to the input first frame, reported separately for Subject~(S) and Background~(B).}
    \label{tab:vbench_grouped}
    \setlength{\tabcolsep}{3pt}
    \resizebox{\columnwidth}{!}{%
    \begin{tabular}{l ccccc}
      \toprule
      Metric & TravelUAV \cite{traveluav} & IndoorUAV \cite{indooruav} & OpenFly \cite{openfly} & UAV-Flow \cite{uavflow} & \textbf{Ours} \\
      \midrule
      Aesthetic
        & $0.471 \pm 0.057$
        & $0.546 \pm 0.053$
        & \underline{$0.551 \pm 0.052$}  
        & $0.549 \pm 0.047$
        & $\mathbf{0.586 \pm 0.037}$ \\
      Smoothness
        & $0.888 \pm 0.026$
        & $0.897 \pm 0.021$
        & $0.927 \pm 0.022$
        & $\mathbf{0.987 \pm 0.006}$
        & \underline{$0.952 \pm 0.016$} \\ 
      Sem.\ Cons.
        & $0.710 \pm 0.048$
        & $0.858 \pm 0.056$
        & $0.883 \pm 0.037$
        & $\mathbf{0.979 \pm 0.008}$
        & \underline{$0.952 \pm 0.013$} \\ 
      I2V Cons.\ S
        & $0.776 \pm 0.051$
        & $0.913 \pm 0.044$
        & $0.912 \pm 0.062$
        & $\mathbf{0.985 \pm 0.006}$
        & \underline{$0.977 \pm 0.010$} \\ 
      I2V Cons.\ B
        & $0.797 \pm 0.044$
        & $0.930 \pm 0.029$
        & $0.923 \pm 0.041$
        & $\mathbf{0.974 \pm 0.006}$
        & \underline{$0.973 \pm 0.007$} \\ 
      \midrule
      \textbf{Avg}
        & $0.7284 \pm 0.032$
        & $0.8288 \pm 0.034$
        & $0.8392 \pm 0.032$
        & $\mathbf{0.8948 \pm 0.011}$
        & \underline{$0.888 \pm 0.012$} \\ 
      \bottomrule
    \end{tabular}}
\end{table}
\begin{figure}[h]
  \centering
  \includegraphics[width=0.5\textwidth]{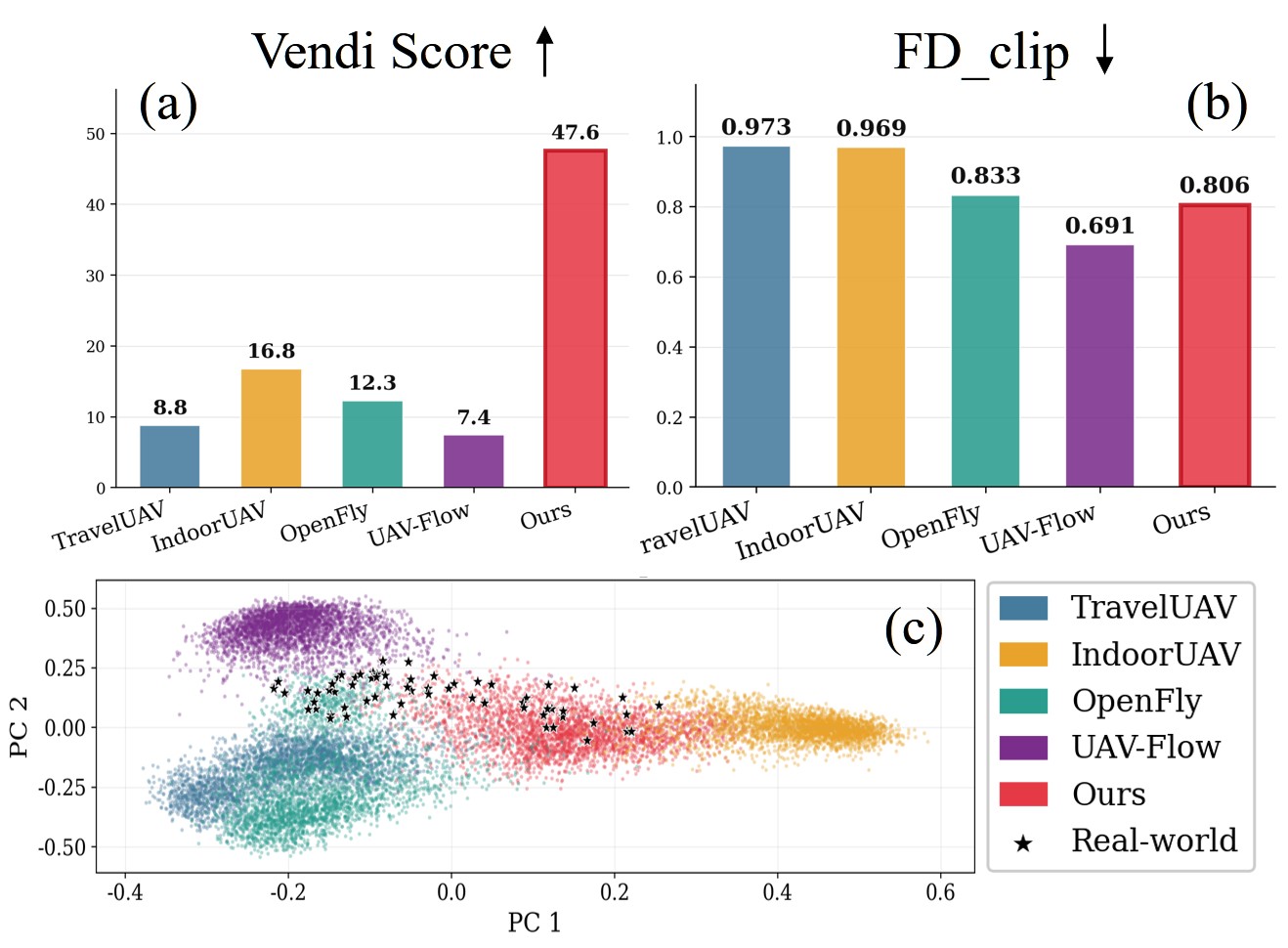}
  \vspace{-5mm}
  \caption{\textbf{Dataset diversity analysis.} Figure (a) and (b) show the Vendi Score and FD CLIP. Figure (c) is the visualization of the clip feature distribution of the datasets.}
  \label{fig:diversity}
  \vspace{-4mm}
  
\end{figure}

\subsection{Datasets Benchmark}

\subsubsection{Datasets Diversity and Sim-to-real Benchmark}
We evaluate our dataset against existing UAV-oriented datasets along two complementary axes: semantic diversity and sim-to-real distribution alignment.
For diversity, we adopt the Vendi Score~\cite{friedman2023vendi}, which quantifies the effective number of semantically distinct modes via the matrix exponential entropy of the CLIP~\cite{radford2021learning} feature similarity matrix.
As shown in Fig.~\ref{fig:diversity}, our dataset achieves a Vendi Score of 47.6, surpassing the next highest competitor IndoorUAV (16.8) by a factor of $2.8\times$, indicating substantially richer coverage across scene types, lighting conditions, and viewpoints.
For sim-to-real alignment, we compute the Fréchet Distance~\cite{heusel2017gans} over CLIP features (FD$\text{CLIP}$) between each dataset and a held-out set of 1000 real-world deployment images.
Our dataset achieves FD$\text{CLIP}$ = 0.806, the lowest among all non-real-world datasets, and approaching the score of the real-flight dataset UAV-Flow (0.691).
The Principal Component Analysis (PCA) projection in Fig.~\ref{fig:diversity} further confirms this qualitatively: real-world samples ($\bigstar$) are visually embedded within our dataset's distribution, while competing datasets occupy largely disjoint regions of the feature space. Together, these results demonstrate that our dataset uniquely combines broad semantic diversity with strong real-world distributional alignment.

\subsubsection{Video Model Trained With Different Datasets Benchmark}

\label{nav_evaluation}

\paragraph{Visual Quality Evaluation}
To validate that our dataset can bridge the sim-to-real gap and generalize to unseen real-world environments, we fine-tune a video generation model, Wan2.2 5B, separately on each of the five datasets. Since the datasets have different sizes and each dataset contains variable-length videos, we adopt an autoregressive training paradigm and group videos with similar lengths into separate training buckets before training \cite{helios}. We collect 1000 real-world text-image pairs as an evaluation dataset and first evaluate the resulting checkpoints using VBench~\cite{vbench}.

As shown in Table~\ref{tab:vbench_grouped}, our dataset achieves the highest VBench average among all non-real-world datasets at $0.866$, outperforming IndoorUAV ($0.806$), OpenFly ($0.820$), and TravelUAV ($0.714$).  Notably, ours closely approaches UAV-Flow ($0.874$), which is a purely real-world aerial dataset. This result suggests that our dataset contains sufficient visual realism to produce generated videos of perceptual quality.

\paragraph{Navigation Ability Evaluation}

While VBench measures visual quality, it cannot assess navigation capabilities. Unlike previous simulation-based benchmarks (e.g., R2R \cite{r2r}) that suffer from sim-to-real gaps, we propose a real-world navigation benchmark with 1,000 authentic text-image pairs. Leveraging the world action model paradigm, we generate videos to predict task success rates, which closely align with real-world execution as validated in Section~\ref{short_horizon}. Following recent practices \cite{liu2026improving}, we employ Qwen3.6-27B to automatically assess navigation performance using a four-level ordinal rubric:

\begin{itemize}
  \item \textbf{L0}: No usable navigation.
  \item \textbf{L1}: Usable navigation, but the prompted motion is not fully followed.
  \item \textbf{L2}: The prompted motion is followed, but the target view is not reached.
  \item \textbf{L3}: The prompted motion is followed and the target view is reached.
\end{itemize}

We compute the average navigation score by mapping L0--L3 to integer values 0--3 and averaging over the 1000 videos. To validate the reliability of this automatic scoring pipeline, we compare its outputs against ratings from a human expert on the same set of videos. The pipeline's score exactly matches the human expert's rating on 56.9\% of videos. An additional 32.8\% differ by exactly one level, bringing the total agreement within one level to 89.7\%. These results provide an initial assessment of the automatic evaluator’s agreement with human ratings.

\begin{figure}[h]
  \centering
  \includegraphics[width=0.5\textwidth]{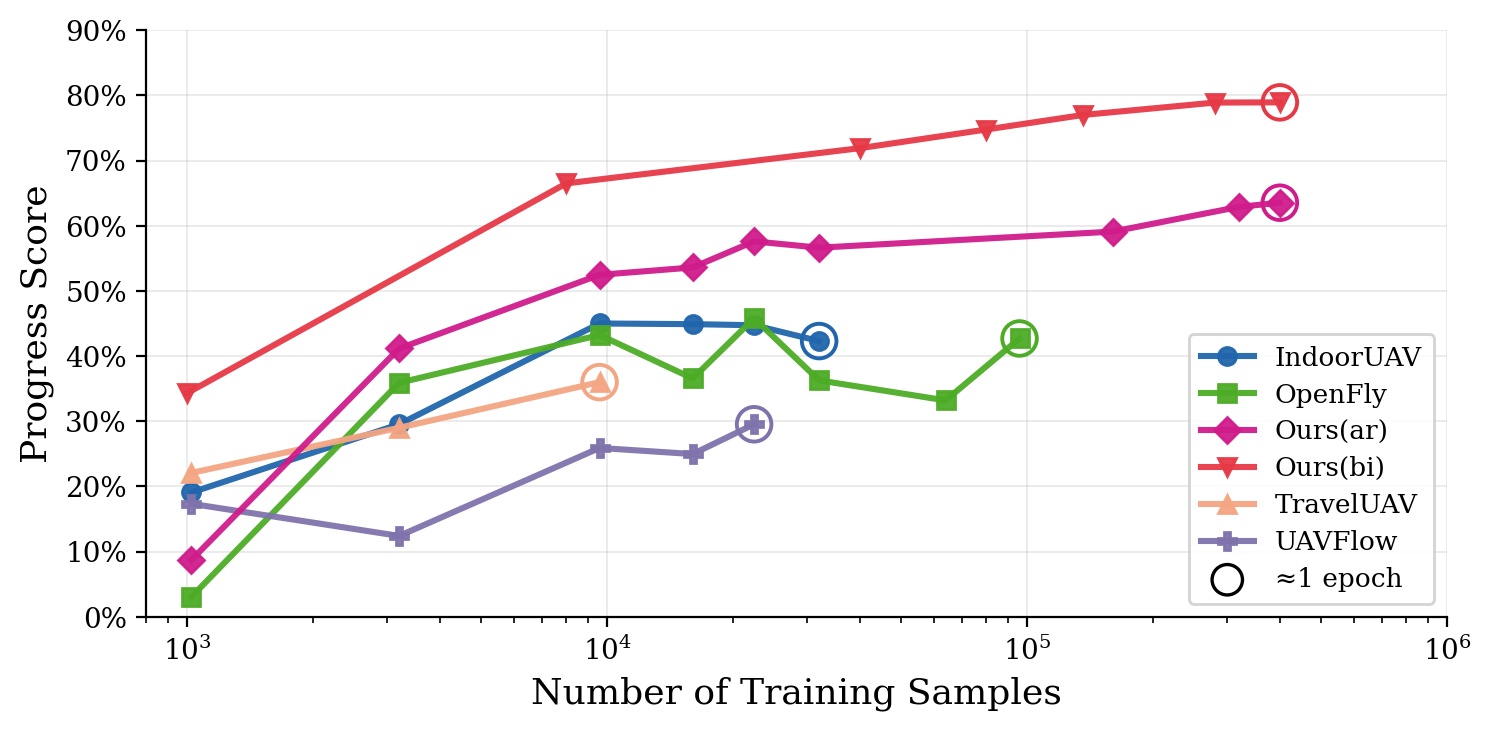}
  \vspace{-7mm}
  \caption{\textbf{Navigation performance scaling with dataset size.} The ar and bi mean autoregressive and bidirectional, respectively.}
  \vspace{-5mm}
  
  \label{fig:scaling_score}
\end{figure}

As shown in Fig.~\ref{fig:scaling_score}, training on different datasets produces very different results. The progress score is the average over the four levels. UAV-Flow stays below 30\%, as its real-flight data are collected only on a university campus with limited diversity. TravelUAV reaches only around 36\% due to the severe sim-to-real gap. IndoorUAV improves quickly at first but saturates at around 45\% within the first epoch. OpenFly is the largest comparison dataset with 100k videos, but its homogeneous data makes performance fluctuate between 33\% and 46\% with no clear gain from more training. In contrast, our dataset improves steadily under both paradigms, reaching around 63\% (autoregressive) and 79\% (bidirectional) at one epoch, far above all comparison datasets. This shows that our pipeline can generate large-scale, high-fidelity, and diverse datasets that generally improve embodied navigation capabilities.

\begin{figure}[h]
  \centering
  \includegraphics[width=0.5\textwidth]{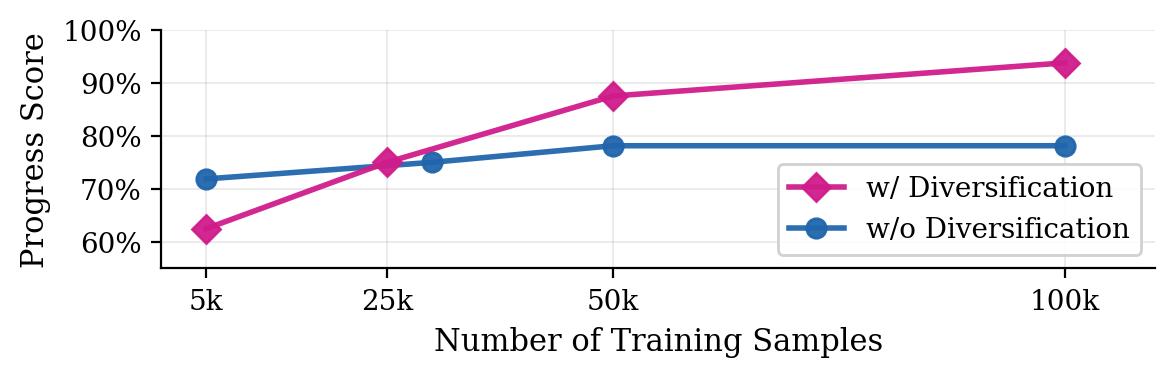}
  \vspace{-5mm}
  
  \caption{\textbf{Ablation study on style diversification.} Performance comparison between training on original 10K samples versus 100K diversified samples on unseen real-world tasks.}
  \label{fig:exp_diversification}
  \vspace{-2mm}
\end{figure}

\subsection{Ablation Study}

\subsubsection{Motion Rebalancing}

We conduct an ablation study under two settings: training on the
original data without motion rebalancing and training on the
motion-rebalanced data. We train separate models under identical
conditions and evaluate the generated videos using the evaluation
pipeline described above. The results show that motion rebalancing
improves the navigation success rate by approximately 15\%, showing its effectiveness in reducing motion-distribution bias and improving policy performance.

\subsubsection{Style Diversification}
To validate the effectiveness of style diversification, we train on the diversified 100K dataset versus the original 10K dataset. We assess generalization on 100 unseen real-world test tasks. As shown in Fig.~ \ref{fig:exp_diversification}, training without diversification plateaus at 78.1\% progress score after 50K samples and gains nothing further. With diversification, performance continues to climb, reaching 93.75\% at 100K samples, a 15.6-point improvement. The result shows that the original 10K data is simply too narrow for the model to generalize. Diversification breaks this ceiling by broadening the training distribution.

\subsection{Real-World Experiment}
To evaluate whether our dataset can alleviate the sim-to-real gap and generalize to unseen environments, we conduct several experiments.

\begin{table}[t]
\centering
\caption{Inference time of each stage in our pipeline.}
\label{tab:inference_time}
\resizebox{\columnwidth}{!}{%
\begin{tabular}{lcccc}
\toprule
Stage & Prompt rewrite & Video generation & Action decode & Ego-planner \\
\midrule
Time  & 1 s & 0.7 s & 0.5 s & 10 ms \\
\bottomrule
\end{tabular}%
}
\vspace{-3mm}

\end{table}

\subsubsection{Real-World Deployment}
We run video model inference on one A100 server. The VLM, Qwen3.6-27B, is deployed on two A100 servers and is used for prompt rewriting. Real-time frames are sent to the server via TCP to maintain a history frame buffer. When a human command is triggered, the VLM rewrites the command into the style of the training datasets to reduce prompt distribution mismatch. The rewritten prompt is then sent to the video model to generate a full-pixel video, which is decoded into a list of waypoints by our action decoder.
The drone platform is equipped with a DJI Osmo Action 5 Pro for RGB observations and a LiDAR for pose estimation. The decoded discrete waypoints are used to initialize a trajectory, and Ego-Planner \cite{ego} optimizes this trajectory using collision information to achieve collision-free execution. The total inference time is shown in Table \ref{tab:inference_time}.

\begin{figure}[h]
  \centering
  \includegraphics[width=0.5\textwidth, height=9cm]{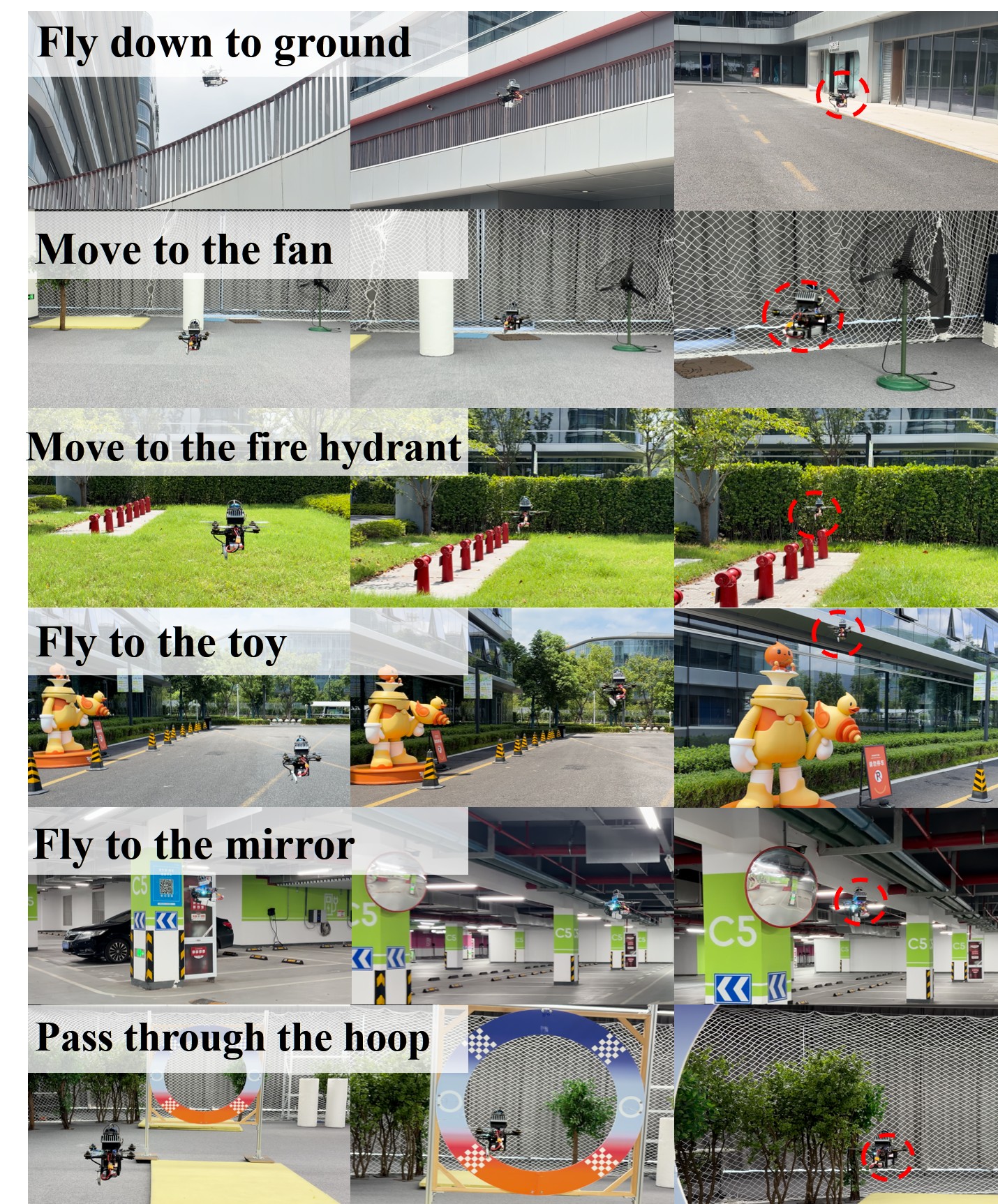}
  
  \caption{\textbf{Short-horizon navigation experiments in real-world environments.}}
  
  \label{fig:shot_distance}
  \vspace{-2mm}
\end{figure}

\begin{figure}[h]
  \centering
  \includegraphics[width=0.5\textwidth, height=8cm]{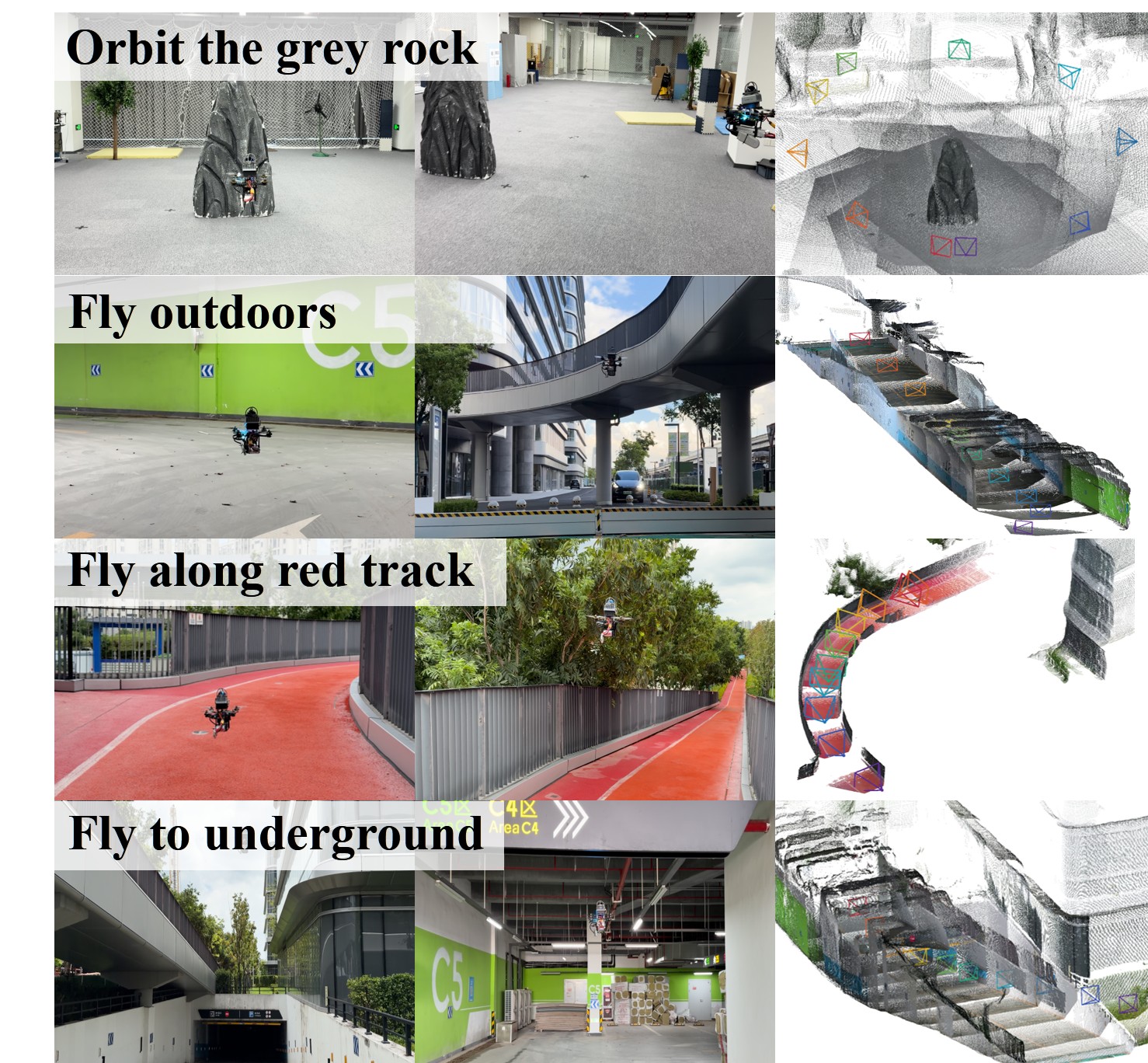}
  \vspace{-5mm}
  \caption{\textbf{Long-horizon navigation demonstration.}}
  \vspace{-5mm}
  
  \label{fig:long_distance}
\end{figure}

\subsubsection{Short Horizon Experiment}
\label{short_horizon}

We evaluate our method on 20 real-world tasks (5 trials each), achieving a 75\% overall success rate. Failures primarily occur when targets are too small or lack color contrast with the background. As shown in Fig.\ref{fig:shot_distance}, these trials span diverse indoor and outdoor scenes, demonstrating that our data pipeline effectively bridges the sim-to-real gap and generalizes well.

\subsubsection{Long Horizon Experiment}

We evaluate our long-video generation pipeline on long-horizon navigation tasks. As shown in Fig.~\ref{fig:long_distance}, some tasks exceed the length of a single clip. We address this by using previously observed RGB frames retrieved from the onboard observation buffer using odometry as context to resume generation, ensuring new segments inherit the preceding state and maintain motion continuity. Notably, the success rate remains stable when the target is clearly visible in the first frame, but drops to 50\% when the target is ambiguous. These results show our pipeline effectively handles long-horizon tasks with a reasonable success rate under the bidirectional training paradigm.

\section{CONCLUSIONS}

In this paper, we propose NavGen, an embodied 3D navigation dataset generation pipeline based on visual generative models. We generate 300K diverse navigation episodes using a scalable open-source video model, and produce an additional 10K high-quality long-tail samples with a more powerful but closed-source model, which are then expanded to 100K through style diversification, yielding 400K episodes in total. Motion rebalancing is applied throughout to correct distribution biases in the generated data. Benchmark comparisons against existing embodied 3D navigation datasets show that models trained on our data consistently outperform those trained on prior datasets on both video-quality metrics and navigation success rate. To validate real-world transferability, we carry out zero-shot deployment across diverse indoor and outdoor environments, achieving a task success rate of approximately 75\%. Together, these results demonstrate that large visual generative models are effective scalable data engines for embodied navigation.

\bibliographystyle{IEEEtran}
\bibliography{IEEEabrv,ref}

\end{document}